\documentclass[preprint,12pt]{elsarticle}

\usepackage{amssymb}
\usepackage{hyperref}
\usepackage{multirow,tabularx}
\usepackage{adjustbox}
\usepackage{textcomp}
\usepackage{etoolbox,lipsum,cleveref}
\usepackage[font=footnotesize,labelfont=bf]{caption}
\usepackage[font=footnotesize,labelfont=bf]{subcaption}

\begin{document}

\begin{frontmatter}



\title{TNCR: Table Net Detection and Classification Dataset}

\author[inst1,inst2]{Abdelrahman Abdallah}

\author[inst1,inst2]{Alexander Berendeyev}
\author[inst1,inst2]{Islam Nuradin}
\author[inst1,inst2]{Daniyar Nurseitov}
\affiliation[inst1]{organization={ Department of Machine Learning \& Data Science },
            addressline={Satbayev University}, 
            city={Almaty},
            postcode={050013}, 
            state={Almaty},
            country={Kazakhstan}}
\affiliation[inst2]{organization={National Open Research Laboratory for Information and Space Technologies},
            addressline={Satbayev University}, 
            city={Almaty},
            postcode={050013}, 
            state={Almaty},
            country={Kazakhstan}}

\begin{abstract}
We present TNCR, a new table dataset with varying image quality collected from free websites. TNCR dataset can be used for table detection in scanned document images and their classification into 5 different classes. TNCR contains 9428 high-quality labeled images. In this paper, we have implemented state-of-the-art deep learning-based methods for table detection to create several strong baselines. Cascade Mask R-CNN  with ResNeXt-101-64x4d Backbone Network achieves the highest performance compared to other methods with a precision of 79.7\%, recall of 89.8\%, and f1 score of 84.4\% on the TNCR dataset. We have made TNCR open source in the hope of encouraging more deep learning approaches to table detection, classification and structure recognition. The dataset and trained model checkpoints are available at \url{https://github.com/abdoelsayed2016/TNCR_Dataset}
\end{abstract}


\end{frontmatter}


\section{Introduction}
\label{sec:sample1}
With so many applications, tools, and online platforms booming in today's technological era, the amount of data being collected is rapidly increasing. To effectively handle and access this massive amount of data, valuable information extraction tools must be developed. The fetching and accessing of data from tabular forms is one of the sub-areas in the Information Extraction field that requires attention. Several industries around the world, particularly the banking and insurance industries, rely heavily on paperwork and documentation. Tables are commonly utilized for anything from recording client information to reacting to their requirements. This information is then sent as a document (hard copy) to other departments for approval, where miscommunication can occasionally result in problems when grabbing data from tables. Instead, we can directly scan such documents into tables and work on the digitized data once the original data has been acquired and authorized.

\begin{figure}[h!]
    \begin{subfigure}{0.49\textwidth}
        \includegraphics[width=\textwidth,height=7cm]{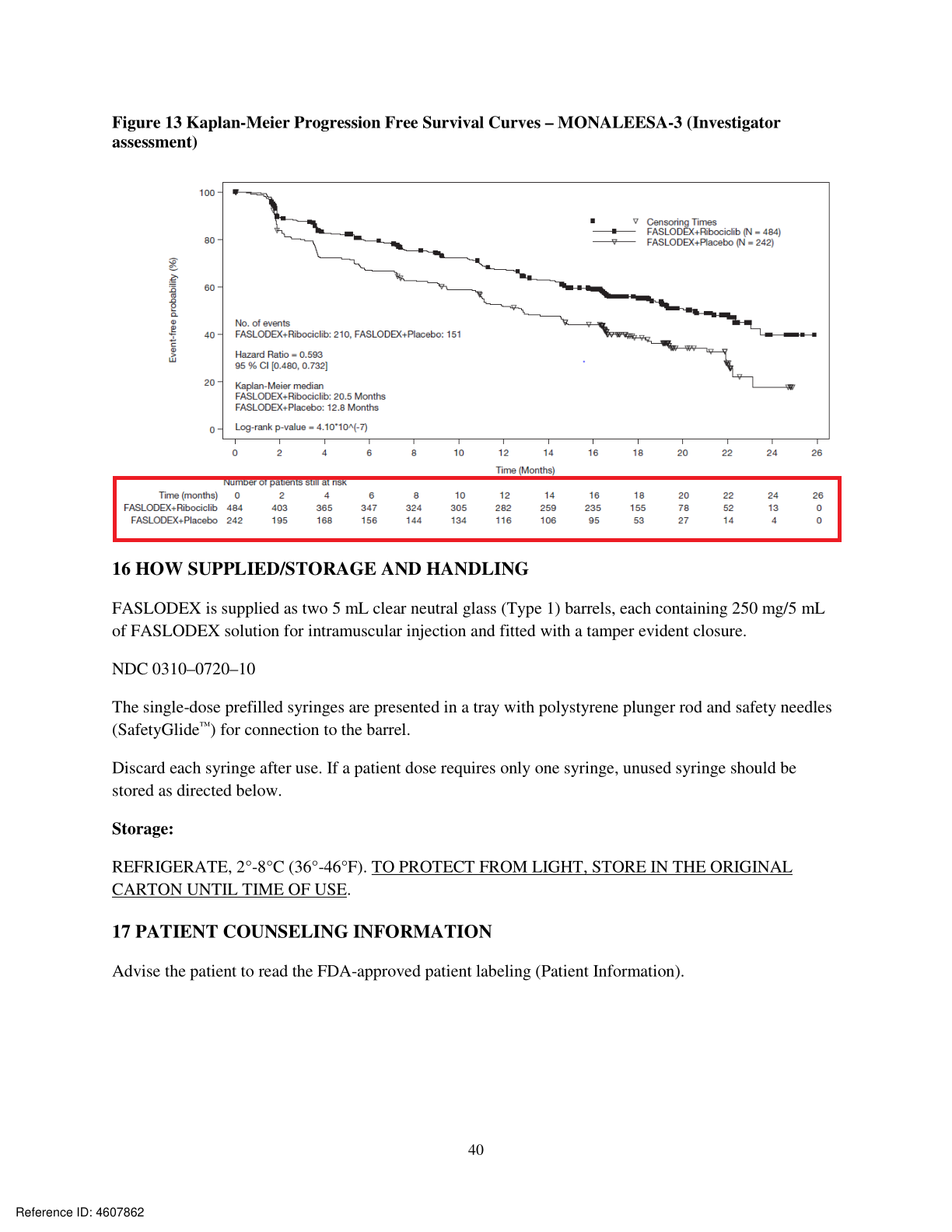}
        \caption{}
        \label{fig:example1}
    \end{subfigure}
    \begin{subfigure}{0.49\textwidth}
        \includegraphics[width=\textwidth,height=7cm]{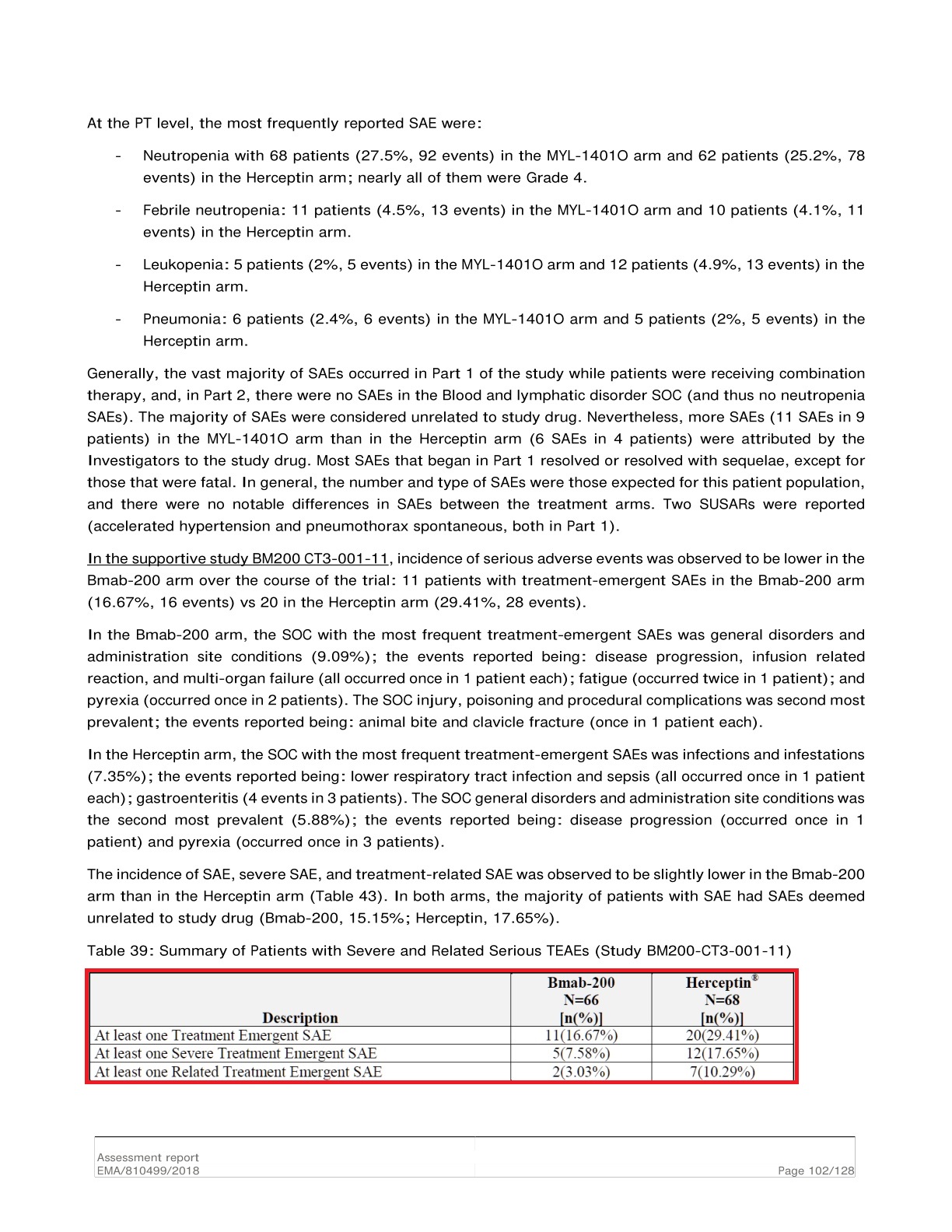}
        \caption{}
        \label{fig:example2}
    \end{subfigure}
    \begin{subfigure}{0.49\textwidth}
        \includegraphics[width=\textwidth,height=7cm]{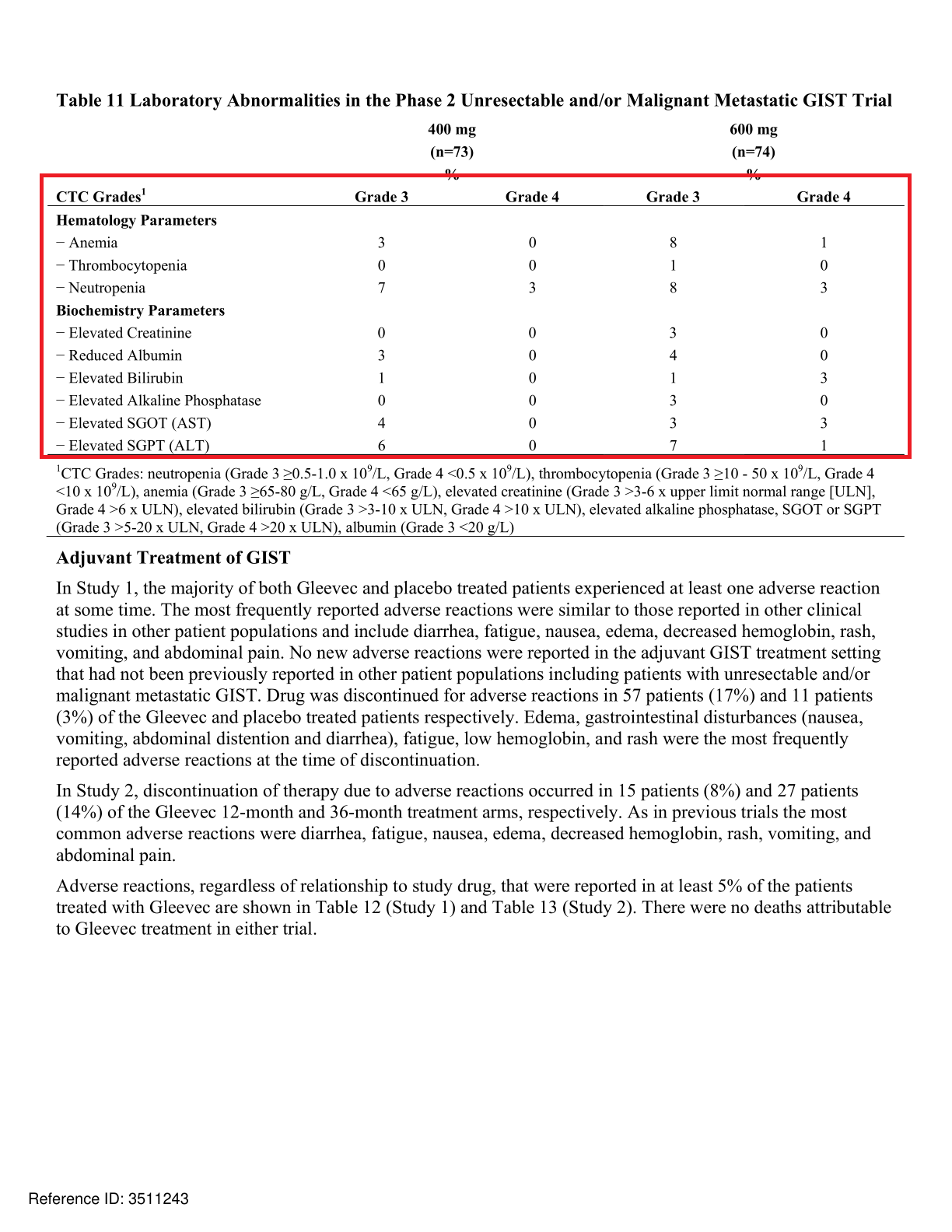}
        \caption{}
        \label{fig:example3}
    \end{subfigure}
    \begin{subfigure}{0.49\textwidth}
        \includegraphics[width=\textwidth,height=7cm]{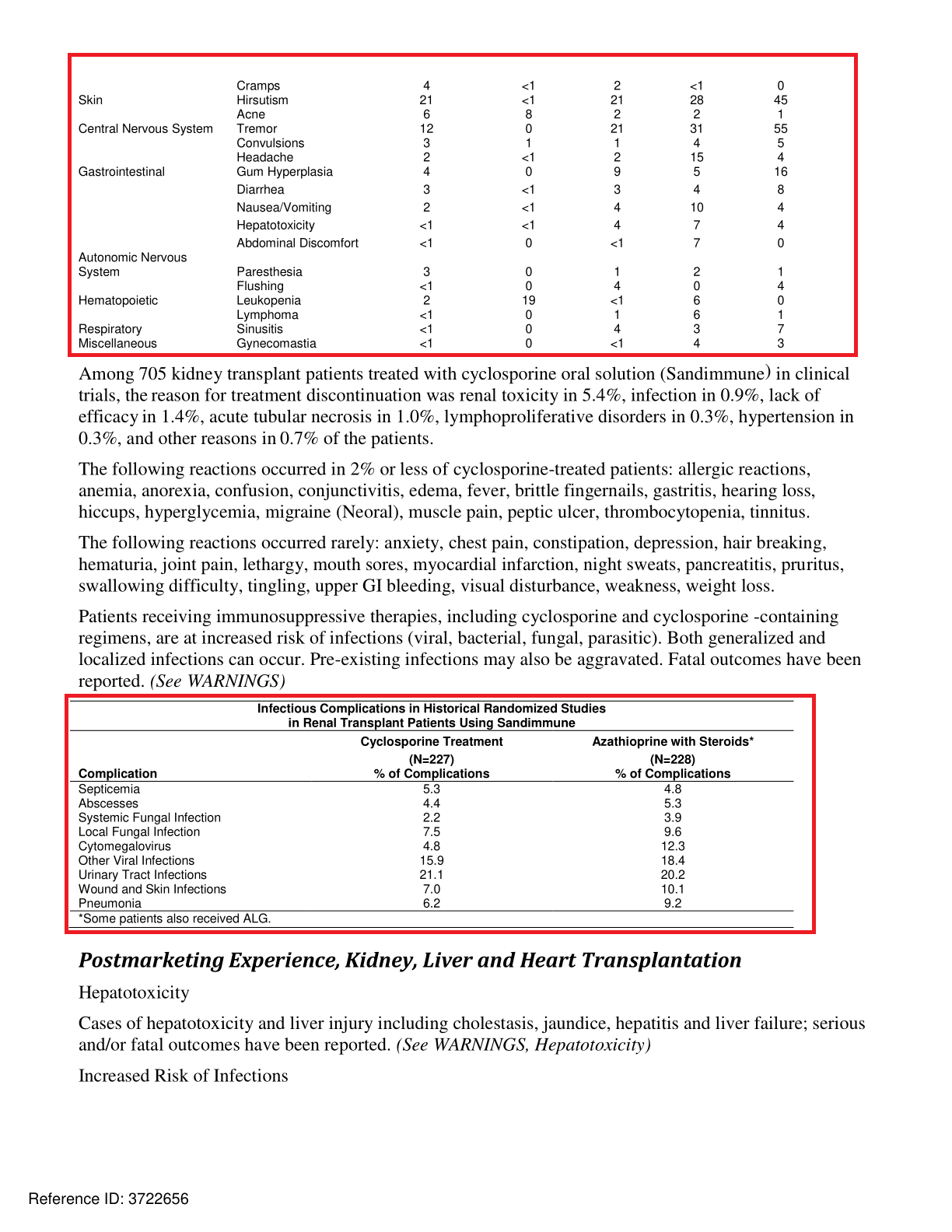}
        \caption{}
        \label{fig:example4}
    \end{subfigure}
\caption{Electronic image examples in various formats and layouts from our dataset} 
\label{fig:examples_images}
\end{figure}

Table detection and structure recognition is an essential task in images analysis for automatically extracting information from the table in a digital way. image or document table detection and extraction is difficult because of the format of the document and various table layouts as shown in Fig. \ref{fig:examples_images}. Recently, deep learning had a significant impact on computer vision specially on image-based approaches for table detection, information extraction and analysis. A few studies have been conducted on the identification of tables in documents \cite{8270123,traquair2019deep,gilani2017table,tran2015table,7490132}. However, there is significantly less work put into detecting table structures, and the table structure is frequently classified by the rows and columns of a table \cite{mao2003document,kara2020holistic,sarkar2019document}.

Deep learning has recently achieving state-of-the-art using convolutional neural network (CNN) \cite{lecun1995convolutional} in many tasks including object detection \cite{zhao2019object}, face recognition \cite{lawrence1997face}, sequence to sequence learning \cite{gehring2017convolutional,Abdallah_Abdelrahman}, speech recognition \cite{abdel2014convolutional}, semantic segmentation \cite{paszke2016enet}, image classification \cite{li2014medical}, handwritten recognition \cite{Abdallah_2020,nurseitov2020hkr,Daniyar_2020}, and table detection \cite{8270123,sarkar2019document,mao2003document} is demanding because they need to classify tables among the texts and other figures. The presence of split columns or rows, as well as nested tables or embedded figures, makes the detection of a table even more difficult.

In this paper, we propose a new dataset called Table Net Detection and Classification Dataset (TNCR) that can be used for table detection and classification of tables into 5 different class. Also, we train deep learning models to solve the two tasks and compare them. Table detection  is performed by using instance segmentation on each image. Each instance of the segmented table detects at pixel level at the images. In addition, we used same model for classifying the segmented tables into 5 different classes.


The main contribution of our research are summarized as follows:
\begin{itemize}
    \item  First, this work presents a new dataset for table detection and table classification. It contains images of different quality for training and testing. The images are real, not generated from LATEX or Word documents. Our dataset contains 9428 images with 5 different labels for table classification (Full lined, No Lines, Merged cells, Partial lined, Partial line merged cells).
    \item 	Second, we present a brief description of deep learning models for object detection and classification that and present comparative results. For a better understanding of models performance, COCO performance metrics over IoUs ranging from 50\% to 95\% are displayed for each model.
    \item Third, we built many robust baselines using state-of-the-art models with end-to-end deep neural networks to test the effectiveness of our dataset. we compared state-of-the-art object detection models like  Cascade R-CNN \cite{cai2019cascade}, Cascade mask R-CNN \cite{cai2019cascade}, Cascade RPN \cite{vu2019cascade}, Hybrid Task Cascade\cite{chen2019hybrid}, and  YOLO \cite{redmon2018yolov3} with different backbone combinations presented as follow ResNet-50 \cite{he2016deep},  ResNet-101 \cite{he2016deep} and ResNeXt101 \cite{xie2017aggregated}. some models are trained in different learning schedule (1x, 20e and 2x).
\end{itemize}

The rest of paper is structured as follows: Section \ref{relatedwork} presents the related work on the topics of existing datasets and a brief history of the methods used in machine learning and deep learning on table detection and structure detection. Section \ref{dataset} describes our dataset in table detection and classification. Section \ref{sec:methodology} provides details description of the models and methodology in object detection (TNCR). Section \ref{Experiments_Results} presents experimental results with a comprehensive analysis of table detection using different models and summary of the paper and the future work are described in Section \ref{Conclusion}.


\section{Related Work}
\label{relatedwork}
\subsection{Existing Datasets}

ICDAR2013 dataset \cite{gobel2013icdar} contains 150 tables, with 75 tables in 27 EU excerpts and 75 tables in 40 US Government excerpts. Table regions are rectangular areas of a page that are defined by their coordinates. Because a table can span multiple pages, multiple regions can be included in the same table. ICDAR2013 is split up into two sub-tasks, table detection or location and table structure recognition. The goal of the table structure recognition task is to compare methods for determining table cell structure given accurate location information. 


UNLV Table dataset \cite{shahab2010open} consists of 2889 pages of scanned document images collected from various sources (Magazines, News papers, Business Letter, Annual Report etc). The scanned images are available in bitonal, greyscale, and fax formats, with resolutions of 200 and 300 DPI. Along with the original dataset, which contains manually marked zones, there is ground truth data; zone types are provided in text format.


The Marmot dataset \cite{fang2012dataset} ground-truths were extracted using the semi-automatic ground-truthing tool ”Marmot” from a total of 2000 pages in PDF format. The dataset is made up of roughly 1:1 ratios of Chinese and English pages. The Chinese pages were chosen from over 120 e-Books from the Founder Apabi library’s diverse subject areas, with no more than 15 pages chosen from each book. The Citeseer website was used to crawl the English pages.


DeepFigures dataset \cite{siegel2018extracting} contains documents with tables and figures from arXiv.com and the PubMed database. The DeepFigures dataset is focused on large-scale table/figure detection and cannot be used for table structure recognition.


TableBank dataset \cite{li2020tablebank} is a new dataset for table detection and structure detection which consists of 417K high-quality labeled tables in a variety of domains, as well as their original documents.


ICDAR2019 \cite{dejean_herve_2019_2649217} proposed a dataset for table detection (TRACK A) and table recognition (TRACK B). The dataset is divided to two types, historical and modern dataset. It contains 1600 images for training and 839 images for testing. Historical type contains 1200 images in track A and B for training and 499 images for testing. Modern type contains 600 images in track A and B for training and 340 images for testing.

\subsection{Table detection and structure detection}

The goal of table detection is to locate tables in a document using bounding boxes and the goal of table structure recognition is to determine a table’s row and column layout information. Table detection has been studied since the early 1990s. Katsuhiko \cite{itonori1993table} explains how to recognize table structure from document images using a new method. Each cell in a table is represented by a row and column pair that is arranged regularly in two dimensions. It coordinates explicitly found even when some ruled lines are missing. As a result, he has assumed that the table structure is defined by an arrangement of tentblocks, which is an arrangement of rows and columns, with ruled lines indicating their relationship. This procedure consists of two steps: expanding the bounding boxes of the cells and assigning row and column numbers to each edge. Wonkyo Seo et al,\cite{seo2015junction} proposes novel junction detection and labeling approaches to increase accuracy, where junction detection involves finding candidates for cell corners and junction labeling implies inferring their connections. Chandra and Kasturi \cite{chandran1993structural} proposed for structure table detection, The document is scanned in order to extract all horizontal and vertical lines. These lines are used to approximate the table’s dimensions. Thomas and Dengel \cite{inproceedings} proposes a novel method for recognizing table structures and analyzing layouts. The analysis of the detected layout components is based on the creation of a tile structure, which reliably recognizes row- and/or column spanning cells as well as sparse tables. The whole method is domain agnostic, may ignore textual contents if desired, and can therefore be used to any mixed-mode document (with or without tables) in any language, and even works with low-quality OCR documents (e.g. facsimiles). All horizontal and vertical lines that are present should be removed. These lines are used to approximate the table’s dimensions.

The rapid development of machine learning in computer vision has had a significant impact on data-driven image-based table detection approaches in 1998 lead Kieninger and Dengel \cite{inproceedings} proposed first unsupervised machine learning method for table detection task. In 2002 Cesarini Francesca et al. \cite{cesarini2002trainable} proposed a supervised machine learning algorithm based on hierarchical representation using the MXY tree. The presence of a table is inferred by looking for parallel lines in the page’s MXY tree. This hypothesis is then supported by the presence of perpendicular lines or white spaces in the area between the parallel lines. Finally, based on proximity and similarity criteria, located tables can be merged. Also machine learning algorithm used for different tasks in table detection and structure detection like using Support vector machine (SVM) for feature extraction proposed by Kasar \cite{6628801} and sequence labeling task by Silva et al \cite{e2009learning}. Silva proposed a hidden Markov models (HMM) for table location by Interdependent classification using probabilistic graphical models. In this paper shows how to incorporate different document structure finders into the HMM. Using machine learning algorithms with table detection lead to improve the accuracy.


Deep learning plays important role in computer vision. Deep learning has a significant impact on scanned image for table detection. For document analysis, convolutional neural networks (CNNs) are the top candidate for deep learning in image processing approaches. CNNs for object detection have been implemented widely in document analysis and image processing \cite{kara2019deep,kara2020holistic,arif2018table,gilani2017table}. Faster-RCNN \cite{ren2015faster} had shown good impact at table detection and achieved state-of-the-art performance on ICDAR-2013. Shoaib et al\cite{8540832}, proposed a method by combining deformable CNN with Faster-RCNN. Deformable convolution bases its receptive field on the input, allowing it to shape its receptive field to match the input. The network can then accommodate tables with any layout to this adaptation of the receptive field.


CascadeTabNet \cite{prasad2020cascadetabnet} is a deep learning-based end-to-end solution that uses a single Convolution Neural Network (CNN) model to solve both table detection and structure recognition problems. CascadeTabNet present a Cascade mask Region-based CNN High-Resolution Network (Cascade mask R-CNN HRNet)-based model that simultaneously detects table regions and classifies detected tables.

DeepDeSRT  \cite{8270123} is contain two steps: first step is deep learning method for table detection where using fine-tuning a pre-trained model of Faster RCNN and second step is deep learning method for table structure recognition by using fine-tuning FCN proposed by Shelhamer et al. \cite{long2015fully}  trained on VOC pascal\cite{everingham2010pascal}.


For both table detection and structure recognition, TableNet  \cite{paliwal2019tablenet} proposed a novel end-to-end deep learning model. To segment out the table and column regions, the model takes advantage of the interdependence between the twin tasks of table detection and table structure recognition. Then, from the identified tabular sub-regions, semantic rule-based row extraction is performed. On the publicly available ICDAR 2013 and Marmot Table datasets, the proposed model and extraction approach were evaluated, yielding state of-the-art results.


Kavasidis et al. \cite{kavasidis2018saliency} proposed a fully convolutional neural network for table and chart detection that overcomes the shortcomings of existing methods. This paper proposes a fully-convolutional neural network based on saliency that performs multi-scale reasoning on visual cues, followed by a fully-connected conditional random field (CRF) for localizing tables and charts in digital/digitized documents.


Leipeng Hao et al. \cite{7490132} proposed a novel method for detecting tables in PDF documents using convolutional neutral networks, one of the most widely used deep learning models. The proposed method begins by selecting some table-like areas using some loose rules, and then building and refining convolutional networks to determine whether the selected areas are tables or not.


\section{Table Net Detection and Classification Dataset (TNCR) }
\label{dataset}
\begin{figure}[h!]
    \begin{subfigure}{0.5\textwidth}
        \includegraphics[width=\linewidth,keepaspectratio]{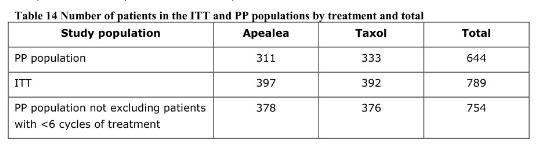}
        \caption{An example for the "Full lined" class}
        \label{fig:Full Line}
    \end{subfigure}
    \begin{subfigure}{0.5\textwidth}
        \includegraphics[width=\linewidth,keepaspectratio]{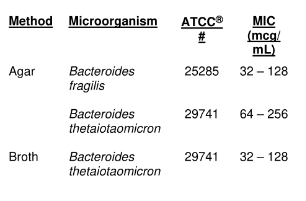}
        \caption{An example for the class "No lines"}
        \label{fig:nolines}
    \end{subfigure}
    \begin{subfigure}{0.5\textwidth}
        \includegraphics[width=\linewidth,keepaspectratio]{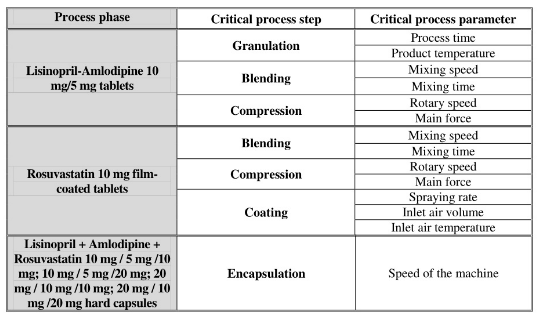}
        \caption{An example for the class "Merged cells"}
        \label{fig:merged_cells}
    \end{subfigure}
    \begin{subfigure}{0.5\textwidth}
        \includegraphics[width=\linewidth,keepaspectratio]{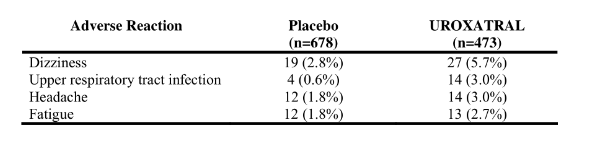}
        \caption{An example for the class "Partial lined"}
        \label{fig:partial_lined}
    \end{subfigure}
    \begin{subfigure}{0.5\textwidth}
        \centering
        \includegraphics[width=\linewidth,keepaspectratio]{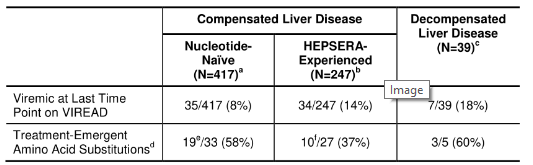}
        \caption{An example for the class "Partial lined"}
        \label{fig:partial_lined_merged_cells}
    \end{subfigure}
\caption{Sample from dataset} 
\label{fig:samples_datase}
\end{figure}

Tables in documents are of different types, they differ from each other in structure or form. The problem for the neural network was a kind of tables, after analyzing all the tables that we have, we classified the tables into 5 groups:
 \begin{enumerate}
     \item Full lined: a table with completely lines, without merged cells (Fig. \ref{fig:Full Line}). Also, Table in which all cells are limited by lines, there are no merged cells and Table in which all columns and rows are delimited by lines on both sides. In this case, the length of all horizontal lines is equal to the width of the table, and the length of the vertical lines is equal to the height.
     \item No lines: a table that has no lines, opposite to the “Full lined” class (Fig. \ref{fig:nolines}).
     \item Merged cells: a table that looks similar to the “Full lined” class, but has at least one merged cell (Fig. \ref{fig:merged_cells}). Merged cell is a full lined , in which two or more cells are concatenated and the contents of the cell are not delimited.
     \item Partial lined: a table that does not have some lines and does not have merged cells (Fig. \ref{fig:partial_lined}). Partial lined is a full lined with one or more lines missing. visually there are pronounced columns, there are no merged cells. column structures are clearly visible, vertical sidelines are absent.
     \item Partial lined merged cells: a table that does not have some lines, but has merged cells (Fig. \ref{fig:partial_lined_merged_cells})
 \end{enumerate}

\begin{figure}[h!]
    \begin{subfigure}{0.5\textwidth}
        \includegraphics[width=\linewidth,keepaspectratio]{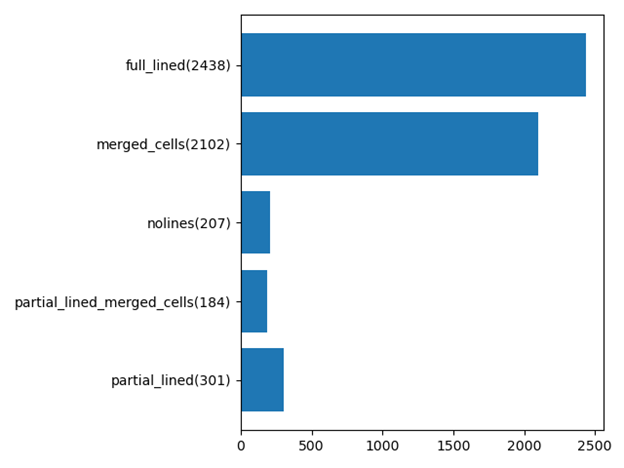}
        \caption{Dataset before re-partitioning}
        \label{fig:histogram_of_dataset_before_balance}
    \end{subfigure}
    \begin{subfigure}{0.5\textwidth}
        \includegraphics[width=\linewidth,keepaspectratio]{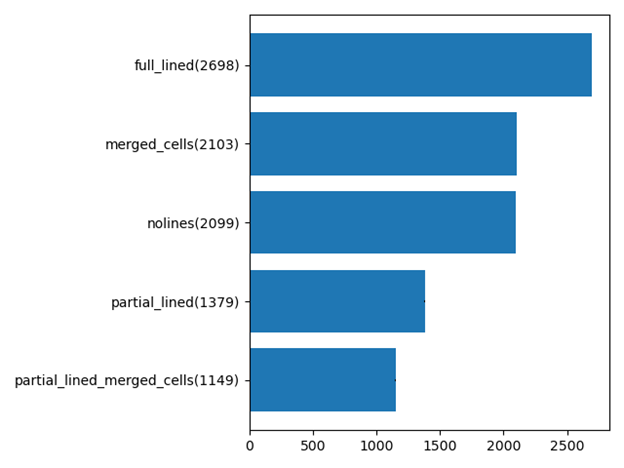}
        \caption{Dataset after re-partitioning}
        \label{fig:histogram_of_dataset_aftere_balance}
    \end{subfigure}
\caption{Histogram of dataset} 
\label{fig:samples}
\end{figure}

In Fig. \ref{fig:histogram_of_dataset_before_balance} show the number of class in the dataset. Since for three classes (No lines, Partial lined merged cells, Partial lined) there were not enough tables for a balanced dataset.
The first model was trained on pure Faster RCNN\cite{ren2015faster} using the luminoth library on the unbalance dataset. It was necessary to find tables in the public domain. And we came to the decision to parse pdf documents from the site accessdata.fda.gov. 875026 pdf pages were parsed, the model recognized 225154 pages with tables. The missing tables for three classes were taken from them and re-partitioned.  Statistics after re-partitioning  shown in Fig. \ref{fig:histogram_of_dataset_aftere_balance}

\section{Methodology}
\label{sec:methodology}
In this section, we describe the methodology of using object detection and classification. We describe different methods and models that we used in table detection and classification.

\subsection{Cascade R-CNN}
\begin{figure}[ht!]
      \centering
      \includegraphics[width=\textwidth,height=.60\textwidth]{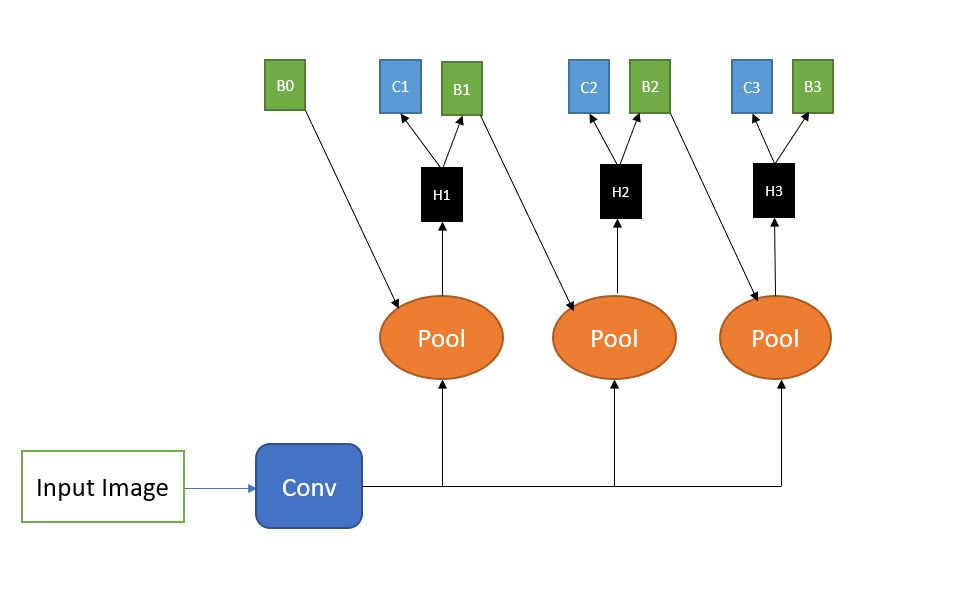}
      \caption{Cascade R-CNN}
      \label{fig:CascadeR-CNN}
\end{figure}

The next problem to address following the R-CNNs is to improve the quality of segmentation and object detection. Quality means making predictions that are more accurate on a pixel level. It is difficult for object detection CNNs to accurately detect objects of various quality and size in an image. This is due to models being trained with a single threshold $u$, which is the Intersection over Union (IoU), being at least 50\% for the object to be considered a positive example. This is quite a low threshold which creates many bad proposals from the Region Proposal Network (RPN) and also makes the networks specialize in making proposals with around $u = 0.50$.
To address this problem, Cai \cite{cai2019cascade} proposed Cascade R-CNN which sets up a multistage network with $u$ increasing at each stage.It uses the same architecture as Faster R-CNN but more of them in a sequence as seen in Fig.  \ref{fig:CascadeR-CNN}. In Faster R-CNN the RPN outputs proposals which are then classified and gets a bounding box. The ones with $u < 0.50$ are discarded. However, instead of being done at this stage, Cascade R-CNN uses the output bounding boxes of the first stage as new region proposals. The second stage increases u and then further refines the output. This is repeated in a third stage and could be repeated as long as memory allows. However, they found that after three stages, the result does not improve further. The key here is that because the network is trained end-to-end, the stages following the initial Faster R-CNN become increasingly better at discarding low-quality proposals of the previous stage. Hence, producing better quality bounding boxes at the final stage.

Fig. \ref{fig:CascadeR-CNN} illustrates the Cascade RCNN architecture. It is a multi-stage extension of the Faster R-CNN architecture.  Cascade RCNN, concentrating on the detection sub-network and using the RPN of the Faster R-CNN architecture for proposal detection. The Cascade R-CNN, on the other hand, isn't limited to this proposal mechanism; other options should be available. 

The first stage is a proposal sub-network, in which a backbone network processes the entire image. like ResNet \cite{he2016deep},To generate preliminary detection hypotheses, known as object proposals, a proposal head (“H0”) is used. A region-of-interest detection sub-network (“H1”), denoted as a detection head, processes these hypotheses in the second stage. Per hypothesis, a final classification score (“C”) and abounding box (“B”) are assigned. Using a multi-task loss with bounding box regression and classification components, the entire detector is learned end-to-end.

\subsection{Cascade Mask R-CNN}
\begin{figure}[ht!]
      \centering
      \includegraphics[width=\textwidth,height=.60\textwidth]{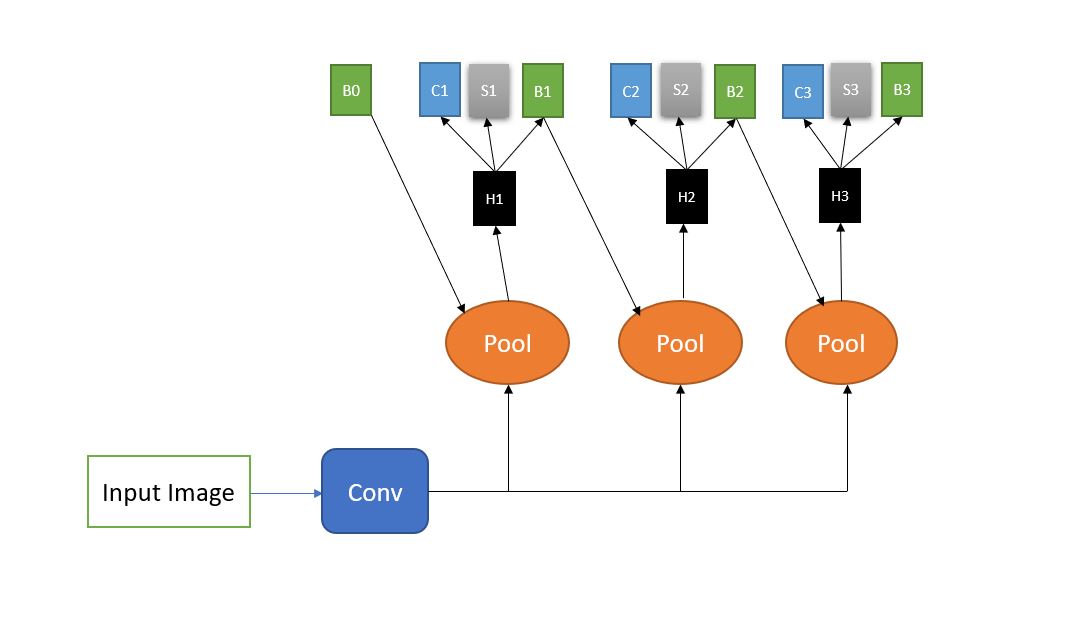}
      \caption{Cascade Mask R-CNN}
      \label{fig:CascadeMaskR-CNN}
\end{figure}

To make it a Cascade Mask R-CNN, it is done similarly as making Faster R-CNN to Mask R-CNN by adding a segmentation branch in parallel to the bounding box regression and classification as seen in Fig. \ref{fig:CascadeMaskR-CNN}.  This is due to segmentation being a pixel-wise operation and is not necessarily improved by having a well-defined bounding box. In the article, they propose using a mask-segmentation branch in the first stage due to being the least computationally heavy. The segmentation branch is added parallel to the detection branch in the Mask R-CNN. The Cascade R-CNN, on the other hand, has several detection branches.

\subsection{Cascade RPN}

\begin{figure}[ht!]
      \centering
      \includegraphics[width=\textwidth,height=.60\textwidth]{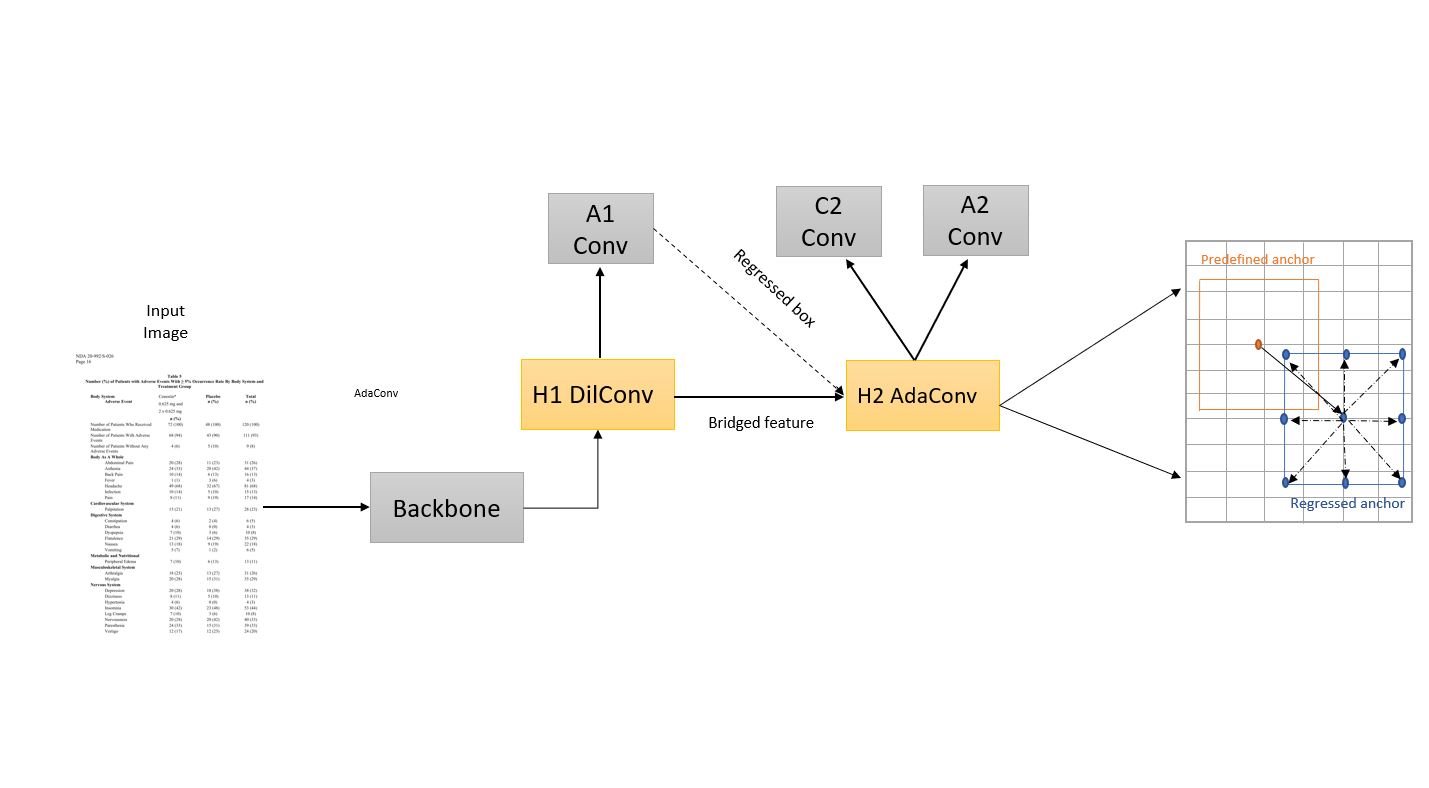}
      \caption{Cascade RPN}
      \label{fig:Cascade RPN}
\end{figure}

Fig. \ref{fig:Cascade RPN} depicts the architecture of a two-stage Cascade RPN\cite{vu2019cascade}. Cascade RPN uses adaptive convolution to align the features to the anchors in this case. Because the anchor center offsets are zeros, the adaptive convolution is set to perform dilated convolution in the first stage. Because the spatial order of the features is maintained by the dilated convolution, the features of the first stage are "bridged" to the next stages.
\subsection{Hybrid Task Cascade (HTC)}
The Hybrid Task Cascade (HTC) \cite{chen2019hybrid} is a new instance segmentation cascade architecture. The main idea is to improve information flow by incorporating cascade and multi-tasking at each stage, as well as leveraging spatial context to improve accuracy even more. HTC designed a cascaded pipeline for progressive refinement in particular.

\begin{figure}[ht!]
      \centering
      \includegraphics[width=\textwidth,height=.60\textwidth]{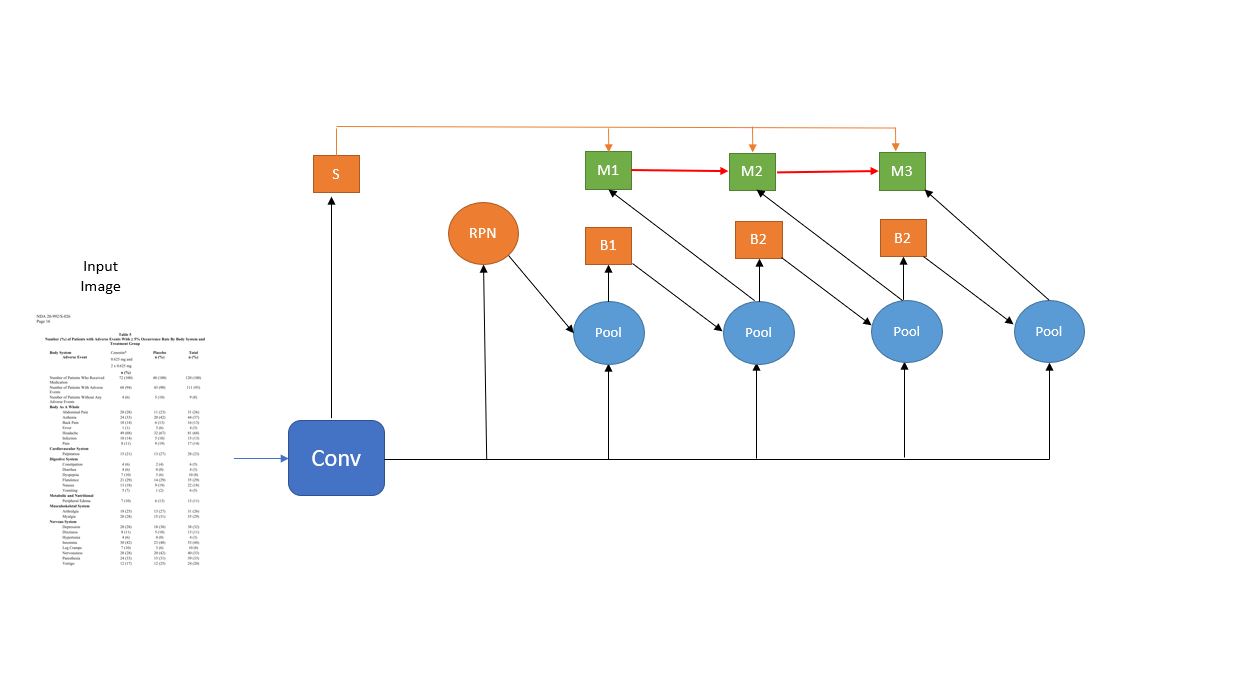}
      \caption{Hybrid Task Cascade}
      \label{fig:Hybrid Task Cascade}
\end{figure}

HTC is a new framework for segmenting instances as seen in Fig. \ref{fig:Hybrid Task Cascade}. It stands out in several ways when compared to other frameworks:
\begin{itemize}
    \item Instead of running bounding box regression and mask prediction in parallel, it interleaves them.
    \item It includes a direct path for reinforcing the information flow between mask branches by feeding the previous stage's mask features to the current one.
    \item By combining an additional semantic segmentation branch with the box and mask branches, it aims to explore more contextual information.
\end{itemize}

\subsection{YOLO}
YOLOv3 \cite{redmon2018yolov3} uses logistic regression to predict the objectness of each bounding box. If the bounding box prior overlaps a ground truth object by a greater amount than any other bounding box prior, this value should be 1. If the bounding box prior isn't the best, but it overlaps a ground truth object by a certain amount, YOLOv3 ignores the prediction. The 0.5 threshold is employed by YOLOv3. For each ground truth object, YOLOv3 only assigns one prior bounding box. There is no loss in coordinate or class predictions if a bounding box prior is not assigned to a ground truth object.

\section{Experiments Results}
\label{Experiments_Results}
\subsection{Dataset and Metrics performance }
TNCR Dataset can serve as basic research on table detection, structure recognition, and table classification. It contains 5 different classes for tables which can help the researchers to detect the table and classify it even with no rows and columns. In this research, we perform preprocessing for tabular cell recognition in TNCR dataset.  The representation of a table in a machine-readable format, where its layout is encoded according to a pre-defined standard, is known as table structure recognition \cite{jiang2021tabcellnet,zhong2019image}. TNCR Dataset is split into three datasets as follows training, validation and testing dataset. We carefully split the dataset from each class in the dataset 70\% for training and 15\% for validation and 15\% for testing as shown in table. \ref{tab:split_dataset}.

\begin{table}[h!]
\caption{training, validation and testing dataset. }
\begin{tabular}{|c|c|c|c|c|c|} 
    \hline
     & Full lined & No Lines & Merged cell & Partial lined  & Partial lined \\ \hline
     Training   & 1888 & 1469 & 1409 &  965   &   804   \\ \hline
     Validation & 405  & 315  &  302 &  207   &    173   \\ \hline
     testing    & 405  & 315  &  302 &  207   &     172  \\ \hline
     Total      & 2698 & 2099 & 2013 & 1379   & 1149     \\ \hline
\end{tabular}
\label{tab:split_dataset}
\end{table}
To evaluate our result for table detection we calculate the average precision (AP) , average recall (AR) and F1-score with the same ways of standard evaluation metrics for COCO dataset on different Intersection Over Union (IoU) threshold. the precision , recall and F1 score calculate as follow :
,

\begin{equation}
    \textrm{Average Precision (AP)} = \frac{\textrm{True Positive (TP)}}{( \textrm{True Positive (TP)}+\textrm{False Positive (FP)} )}
\end{equation}

\begin{equation}
    \textrm{Average Recall (AR)} = \frac{\textrm{True Positive (TP)}}{( \textrm{True Positive (TP)}+\textrm{False Negative (FN) } )}
\end{equation}

\begin{equation}
    \textrm{F1-score} = \frac{ 2 * (\textrm{AP}* \textrm{AR} )}{(\textrm{AP}+\textrm{AR})}
\end{equation}
We define True Positive detection results consistently and use them to compute precision and recall. The table header and all instances should be included in all recognized regions, ensuring that the entire table in the ground truth is captured\cite{luo2021deep}. The area within the bounding box must be free of any noise that would detract from the tabular region's purity. Other elements in a confusion matrix are represented as FP in all models, which stands for "not being a table with bounding boxes," and FN in all models, which stands for "actual tables with incorrect bounding boxes or no bounding boxes." The AP, AR, and F1-score metrics are calculated using confusion metrics. Confusion matrix elements are represented in all models. To compute the evaluation metrics, we used different IoU thresholds for the overlapping area between the result and the ground truth. IoU is used to determine whether a table region has been correctly detected and to measure the overlapping of the detected boxes.
\subsection{Experiment Settings}
The proposed and tested models have all been implemented using the MMdetection library \cite{chen2019mmdetection} for pytorch. MMDetection is a toolbox for object detection that includes a large number of object detection and instance segmentation methods, as well as related components and modules. It gradually develops into a unified platform that encompasses a wide range of popular detection methods and modern modules. The various features of this toolbox are introduced by MMDetection. The experiments were performed on Google Colaboratory platform and  with 3 Tesla V100-SXM GPUs of 16 GB GPU memory and 16 GB of RAM. Also we run on a machine with 2$\times$ ``Intel(R) Xeon(R) E-5-2680'' CPUs and 4$\times$ ``NVIDIA Tesla k20x''.
All the models have been trained and tested with images scaled to a fixed size of 1300$\times$1500 with batch size 16.  SGD is defined as the optimizer with a momentum of 0.9, weight decay of 0.0001, and the learning rate is 0.02. All models utilize the Feature Pyramid Network (FPN) neck.

\subsection{Results}
The evaluation results of table detection for Cascade Mask R-CNN model with different backbones are shown in Table. \ref{tab:Cascade Mask R-CNN}. This table  shows that ResNeXt-101-64x4d backbone has achieves the highest F1 score of 0.844 over 50\%:95\% and maintains the highest F1 score at various IoUs. ResNeXt-101-32x4d backbone also achieves lower performance at IoUs of 95\%, 90\%, and 50\%:95\%. Resnet-101 backbone with 1$\times$ Lr schedule shows lower performance at  IoU of 50\% to 85\%. Benchmarks are frequently assessed at 50\% IoU or a mean average of 50\% to 95\% IoU. As a result, at 50\% IoU, ResNeXt-101-64x4d backbone has the highest precision and recall (0.891 and 0.975, respectively). 

\begin{table}[h!]
\caption{Cascade Mask R-CNN}
\begin{center}
\begin{adjustbox}{width=1\textwidth}
\begin{tabular}{|c|c|c|c|c|c|c|c|c|c|c|c|c|c|}
    \hline
    \multirow{2}{*}{Backbone}& \multirow{2}{*}{Lr schd}& \multirow{2}{*}{} & \multicolumn{11}{|c|}{ IoU } \\
    \cline{4-14}
     & & & 50\% & 55\% & 60\% & 65\% & 70\% & 75\% & 80\% & 85\% & 90\% & 95\% & 50\%:95\% \\
    
    \hline
    \multirow{3}{*}{Resnet-50}  & \multirow{3}{*}{1x} & Precision   & 0.709  & 0.708 & 0.708 & 0.706 & 0.704 & 0.701 & 0.690 & 0.675 & 0.650 & 0.557 &  0.633\\
    & & Recall & 0.778 & 0.777 & 0.776 & 0.775 & 0.774 & 0.770 & 0.760 & 0.747 & 0.725 & 0.647 & 0.713 \\
    & & F1-Score & 0.741 & 0.740 & 0.740 &  0.738 & 0.737 & 0.733 & 0.723 & 0.709 & 0.685 & 0.598 & 0.670 \\

    \hline

    \multirow{3}{*}{Resnet-50}  & \multirow{3}{*}{20e} & Precision   &  0.713 &  0.713 & 0.711 & 0.711  & 0.709 & 0.707 & 0.702 & 0.688 & 0.663  & 0.587 & 0.650 \\
    & & Recall & 0.775 & 0.775 & 0.774 & 0.773 & 0.773 & 0.769 &  0.764 & 0.752 &  0.729 & 0.663 & 0.719 \\
    & & F1-Score & 0.742 & 0.742 & 0.741 & 0.740 & 0.739 & 0.736 & 0.731 & 0.718 & 0.694 & 0.622 & 0.682 \\
    
    \hline
    \multirow{3}{*}{Resnet-101}  & \multirow{3}{*}{1x} & Precision   & 0.701  & 0.699 & 0.699 & 0.698 & 0.696 & 0.692 & 0.684 & 0.673 & 0.653 & 0.570 & 0.635 \\
    & & Recall & 0.776 & 0.776 &  0.775 & 0.774 & 0.773 & 0.768 & 0.757 & 0.75 & 0.731 &  0.659 &  0.718 \\
    & & F1-Score & \textbf{0.736}* & \textbf{0.735}*  & \textbf{0.735}* & \textbf{0.734}*  & \textbf{0.732}*  & \textbf{0.728}* & \textbf{0.718}* & \textbf{0.709}* & 0.689 & 0.611 & 0.673  \\

    \hline
    \multirow{3}{*}{Resnet-101}  & \multirow{3}{*}{20e} & Precision   & 0.803 & 0.802 & 0.799 & 0.796 & 0.788 & 0.781 & 0.766 & 0.734 & 0.674 & 0.468 & 0.636 \\
    & & Recall & 0.968 &  0.967 & 0.964 & 0.961 &  0.953 & 0.945 & 0.931 &  0.903 & 0.849 & 0.669 & 0.819 \\
    & & F1-Score & 0.877 & 0.876 & 0.873 & 0.870 & 0.862 & 0.855 & 0.840 & 0.809 & 0.751 & 0.550 & 0.715 \\

    \hline
    \multirow{3}{*}{ResNeXt-101-32x4d}  & \multirow{3}{*}{1x} 
    & Precision     & 0.761 & 0.760 & 0.751 & 0.740 & 0.735 & 0.728 & 0.696 & 0.665 & 0.591 & 0.383 & 0.572 \\
    & & Recall      & 0.954 & 0.953 & 0.944 & 0.936 & 0.931 & 0.925 & 0.890 & 0.859 & 0.799 & 0.583 & 0.769 \\
    & & F1-Score    & 0.846 & 0.845 & 0.836 & 0.826 & 0.821 & 0.814 & 0.781 & 0.749 & \textbf{0.679}* & \textbf{0.462}* & \textbf{0.656}* \\

    \hline
    \multirow{3}{*}{ResNeXt-101-64x4d}  & \multirow{3}{*}{1x} 
    & Precision   & 0.891 & 0.891 & 0.889 & 0.886 & 0.885 & 0.881 & 0.871 & 0.853 & 0.822 & 0.703 & 0.797\\
    & & Recall    & 0.975 & 0.975 & 0.973 & 0.970 & 0.969 & 0.965 & 0.958 & 0.942 & 0.917 & 0.820 & 0.898 \\
    & & F1-Score  & \textbf{0.931} & \textbf{0.931} & \textbf{0.929} & \textbf{0.926} & \textbf{0.925} & \textbf{0.921} & \textbf{0.912} & \textbf{0.895} & \textbf{0.866} & \textbf{0.757} & \textbf{0.844}\\

    \hline
\end{tabular}
\end{adjustbox}
\end{center}
\label{tab:Cascade Mask R-CNN}
\end{table}
The results are shown in Table. \ref{tab:CascadeR-CNN} for Cascade-RCNN model with with different backbones was proposed by \cite{cai2019cascade} to achieve high F1 score on object detection datasets. ResNeXt-101-64x4d backbone achieves the highest F1 score of 0.841 over 50\%:95\% and maintains the highest F1 score at various IoUs.  Resnet-50 backbone with 1$\times$ Lr schedule achieve lowest performance at various IoUs. Also Resnet-101 backbone with 1$\times$ Lr schedule shows lower performance at  IoU of 65\% to 70\%.  CascadeTabNet proposed by  \cite{prasad2020cascadetabnet} combined by Cascade-Mask-RCNN and High-Resolution Net (HRNet) and achieved a 1.0 F1 score on the ICDAR2013 dataset. The proposed model is from Table. \ref{tab:Cascade Mask R-CNN} and \ref{tab:CascadeR-CNN} shows that  ResNeXt101 led to an improvement over Resnet101 and Resnet50, with a F1-score of 0.931 compared to 0.877 and 0.742 respectively for Cascade-RCNN. 
\begin{table}[h!]
\caption{Cascade R-CNN}
\begin{center}
\begin{adjustbox}{width=1\textwidth}
\begin{tabular}{|c|c|c|c|c|c|c|c|c|c|c|c|c|c|}
    \hline
    \multirow{2}{*}{Backbone}& \multirow{2}{*}{Lr schd}& \multirow{2}{*}{} & \multicolumn{11}{|c|}{ IoU } \\
    \cline{4-14}
     & & & 50\% & 55\% & 60\% & 65\% & 70\% & 75\% & 80\% & 85\% & 90\% & 95\% & 50\%:95\% \\
    
    \hline
    \multirow{3}{*}{Resnet-50}  & \multirow{3}{*}{1x} & Precision   & 0.699 & 0.698 & 0.698 & 0.697 & 0.695 & 0.689 & 0.682 & 0.667 & 0.637 & 0.528 & 0.613 \\
    & & Recall &   0.776 & 0.698 & 0.698 & 0.775 & 0.772 &  0.765 & 0.758 & 0.745 &  0.719 & 0.623 & 0.699 \\
    & & F1-Score & \textbf{0.735}* & \textbf{0.698}* & \textbf{0.698}* &  \textbf{0.733}*  &  \textbf{0.731}* & \textbf{0.725}* & \textbf{0.717}* & \textbf{0.703}* & \textbf{0.675}* & \textbf{0.571}* & \textbf{0.653}* \\
    
    \hline
    \multirow{3}{*}{Resnet-50}  & \multirow{3}{*}{20e} & Precision   & 0.709 & 0.709 & 0.707 & 0.707 & 0.705 & 0.703 & 0.697  & 0.682 & 0.650 & 0.553 & 0.631\\
    & & Recall & 0.776 & 0.776 & 0.774 & 0.773 &  0.771 & 0.767 & 0.762 & 0.751 & 0.721 &  0.640 &  0.708\\
    & & F1-Score & 0.740 & 0.740 & 0.738 & 0.738 & 0.736 & 0.733 & 0.728 & 0.714 & 0.683 & 0.593 & 0.667\\
    
    \hline
    \multirow{3}{*}{Resnet-101}  & \multirow{3}{*}{1x} & Precision   & 0.700 & 0.699 & 0.699 & 0.697 & 0.695 & 0.691 & 0.686 & 0.672 & 0.648 & 0.547 & 0.624 \\
    & & Recall    & 0.776 & 0.776 & 0.776 & 0.774 & 0.771 & 0.766 & 0.761 & 0.750 & 0.727 & 0.636 & 0.706 \\
    & & F1-Score  & 0.736 & 0.735 & 0.735 & \textbf{0.733}* & \textbf{0.731}* & 0.726 & 0.721 & 0.708 & 0.685 & 0.588 & 0.662 \\

    \hline
    \multirow{3}{*}{Resnet-101}  & \multirow{3}{*}{20e} 
    & Precision   & 0.711 & 0.711 & 0.710 & 0.709 & 0.708 & 0.704 & 0.693 & 0.680 & 0.657 & 0.572 & 0.642 \\
    & & Recall    & 0.776 & 0.776 & 0.775 & 0.774 & 0.772 & 0.769 & 0.756 & 0.745 & 0.723 & 0.649 &  0.712 \\
    & & F1-Score  & 0.742 & 0.742 & 0.741 & 0.740 & 0.738 & 0.735 & 0.723 & 0.711 & 0.688 & 0.608 & 0.675\\
    \hline
    
    \multirow{3}{*}{ResNeXt-101-32x4d}  & \multirow{3}{*}{1x} 
    & Precision   & 0.710 & 0.708 & 0.706 & 0.705 & 0.702 & 0.700 & 0.692 & 0.681 & 0.663 & 0.564 & 0.637\\
    & & Recall    & 0.780 & 0.778 & 0.777 & 0.776 & 0.772 & 0.770 & 0.763 & 0.753 & 0.735 & 0.651 &  0.716\\
    & & F1-Score  & 0.743 & 0.741 & 0.739 & 0.738 & 0.735 & 0.733 & 0.725 & 0.715 & 0.697 & 0.604 & 0.674\\
    \hline
    
    \multirow{3}{*}{ResNeXt-101-64x4d}  & \multirow{3}{*}{1x} 
    & Precision   & 0.894 & 0.894 & 0.892 & 0.892 & 0.890 & 0.886 & 0.877 & 0.862 & 0.831 & 0.703 & 0.798 \\
    & & Recall    & 0.971 & 0.971 & 0.970 & 0.959 & 0.967 & 0.963 & 0.954 & 0.943 & 0.914 & 0.810 & 0.891\\
    & & F1-Score  & \textbf{0.930} & \textbf{0.930} & \textbf{0.929} & \textbf{0.924} & \textbf{0.926} & \textbf{0.922} & \textbf{0.913} & \textbf{0.900} & \textbf{0.870} & \textbf{0.752} & \textbf{0.841}\\

    \hline
\end{tabular}
\end{adjustbox}
\end{center}
\label{tab:CascadeR-CNN}
\end{table}

A comprehensive component-wise analysis is performed to demonstrate the effectiveness of Cascade RPN\cite{vu2019cascade}. Different components are omitted to demonstrate the effectiveness of Cascade RPN. Table. \ref{tab:CascadeRPN} shows the results. We adopted Fast R-CNN and Cascade RPN to improve the table detection. The fast R-CNN method achieves f1 score of 0.804 over 50\%:95\% IoU. The fast R-CNN method achieves better performance for table detection compare with CRPN. CRPN achieves f1 score of 0.609 over 50\%:95\% IoU. we have test  Cascade RPN to measure average recall (AR), which is the average of recalls across IoU thresholds from 0.5 to 0.95 with a 0.05 step, is used to assess the quality of region proposals. the AR achieve 0.994 for fast R-CNN and 0.962 for CRPN method over 50\% IoU. 
\begin{table}[h!]
\caption{Cascade RPN}
\begin{center}
\begin{adjustbox}{width=1\textwidth}
\begin{tabular}{|c|c|c|c|c|c|c|c|c|c|c|c|c|c|c|}
    \hline
    \multirow{2}{*}{Method}&\multirow{2}{*}{Backbone}& \multirow{2}{*}{Lr schd}& \multirow{2}{*}{} & \multicolumn{11}{|c|}{ IoU } \\
    \cline{5-15}
    & & & & 50\% & 55\% & 60\% & 65\% & 70\% & 75\% & 80\% & 85\% & 90\% & 95\% & 50\%:95\% \\
    
    \hline
    \multirow{3}{*}{Fast R-CNN}  & \multirow{3}{*}{Resnet-50}  & \multirow{3}{*}{1x} 
    & Precision    & 0.894 & 0.892 & 0.892 & 0.888 & 0.887 & 0.880 & 0.864 & 0.838 & 0.792 & 0.603 & 0.749 \\
    && & Recall    & 0.994 & 0.993 & 0.992 & 0.987 & 0.985 & 0.978 & 0.964 & 0.941 &  0.901 & 0.744 & 0.869 \\
    && & F1-Score  & 0.941 & 0.939 & 0.939 & 0.934 & 0.933 & 0.926 & 0.911 & 0.886 &  0.842 & 0.666 & 0.804\\
    \hline
    \multirow{3}{*}{CRPN}  & \multirow{3}{*}{Resnet-50}  & \multirow{3}{*}{1x} 
     & Precision   & 0.884 & 0.882 & 0.871 & 0.870 & 0.863 & 0.854 & 0.837 & 0.773 & 0.683 & 0.521 & 0.553 \\
    && & Recall    & 0.962 & 0.959 & 0.958 & 0.956 & 0.949 & 0.932 & 0.919 & 0.885 & 0.813 & 0.697 & 0.679 \\
    && & F1-Score  & 0.921 & 0.918 & 0.912 & 0.910 & 0.903 & 0.8912 & 0.876 & 0.825 & 0.742 & 0.596 & 0.609 \\
    \hline
    	
\end{tabular}
\end{adjustbox}
\end{center}
\label{tab:CascadeRPN}
\end{table}

In comparison to other frameworks, Hybrid Task Cascade (HTC) \cite{chen2019hybrid} is unique in several ways: Instead of running bounding box regression and mask prediction in parallel, it interleaves them. It includes a direct path for reinforcing the information flow between mask branches by feeding the previous stage's mask features to the current one. By combining an additional semantic segmentation branch with the box and mask branches, it aims to explore more contextual information. from Table. \ref{tab:HTC}  shows that Resnet-50 backbone with 1$\times$ Lr schedule  has achieves the highest F1 score of 0.840 over 50\%:95\% and maintains the highest F1 score at various IoUs. Resnet-50 backbone with $20e$ Lr schedule achieves the lowest performance over 50\% to 95\% IoUs. Resnet-101 achieve 2.8\% improvement than Resnet-50 with $20e$ Lr schedule over 50\%:95\%. ResNeXt-101-32x4d and ResNeXt-101-64x4d backbones suffer from overfitting through dataset.

\begin{table}[h!]
\caption{Hybrid Task Cascade}
\begin{center}
\begin{adjustbox}{width=1\textwidth}
\begin{tabular}{|c|c|c|c|c|c|c|c|c|c|c|c|c|c|}
    \hline
    \multirow{2}{*}{Backbone}& \multirow{2}{*}{Lr schd}& \multirow{2}{*}{} & \multicolumn{11}{|c|}{ IoU } \\
    \cline{4-14}
    & & & 50\% & 55\% & 60\% & 65\% & 70\% & 75\% & 80\% & 85\% & 90\% & 95\% & 50\%:95\% \\
    
    \hline
    \multirow{3}{*}{Resnet-50}  & \multirow{3}{*}{1x} 
    & Precision   & 0.886 & 0.884 & 0.883 & 0.882 & 0.879 & 0.874 & 0.863 & 0.838 & 0.790 & 0.687 & 0.787 \\
    & & Recall    & 0.993 & 0.991 & 0.991 & 0.990 & 0.986 & 0.980 & 0.968 & 0.947 & 0.906 & 0.809 &  0.901 \\
    & & F1-Score  & \textbf{0.936} & \textbf{0.934} & \textbf{0.933} & \textbf{0.932} & \textbf{0.929} & \textbf{0.923} & \textbf{0.912} & \textbf{0.889} & \textbf{0.844} & \textbf{0.743} & \textbf{0.840}\\

    \hline
    \multirow{3}{*}{Resnet-50}  & \multirow{3}{*}{20e}     
    & Precision   & 0.860 & 0.858 & 0.857 & 0.856 & 0.848 & 0.842 & 0.828 & 0.804 & 0.746 & 0.523 & 0.691 \\
    & & Recall    & 0.989 & 0.987 & 0.986 & 0.985 & 0.975 & 0.969 & 0.955 & 0.929 & 0.872 & 0.696 &  0.843 \\
    & & F1-Score  & \textbf{0.919}* & \textbf{0.917}* & \textbf{0.916}* & \textbf{0.915}* & \textbf{0.907}* & \textbf{0.901}* & \textbf{0.886}* & \textbf{0.861}* & \textbf{0.804}* & \textbf{0.597}* &  \textbf{0.759}*\\

    \hline
    \multirow{3}{*}{Resnet-101}  & \multirow{3}{*}{1x} 
    & Precision   & 0.867 & 0.866 & 0.864 & 0.860 & 0.856 & 0.849 & 0.836 & 0.817 & 0.771 & 0.576 & 0.722 \\
    & & Recall    & 0.992 & 0.991 & 0.989 & 0.983 & 0.977 & 0.970 & 0.957 & 0.940 & 0.902 & 0.741 & 0.867 \\
    & & F1-Score  & 0.925 & 0.924 & 0.922 & 0.917 & 0.912 & 0.905 & 0.892 & 0.874 & 0.831 & 0.648 & 0.787 \\

    \hline
    
    
    \hline
    
\end{tabular}
\end{adjustbox}
\end{center}
\label{tab:HTC}
\end{table}

Table. \ref{tab:YOLO} shows the performance of YOLO for table detection. YOLO shows low-performance overall the other models and it is not suitable for table detection. we trained YOLO with DarkNet-53 backbones with different Scales (320, 416, 608). DarkNet-53 with 320 scale achieve an f1 scale of 0.492. At 95\% has very low performance with 0.042 of f1 score.
\begin{table}[h!]
\caption{YOLO}
\begin{center}
\begin{adjustbox}{width=1\textwidth}
\begin{tabular}{|c|c|c|c|c|c|c|c|c|c|c|c|c|c|}
    \hline
    \multirow{2}{*}{Backbone}& \multirow{2}{*}{Scale}& \multirow{2}{*}{} & \multicolumn{11}{|c|}{ IoU } \\
    \cline{4-14}
     & & & 50\% & 55\% & 60\% & 65\% & 70\% & 75\% & 80\% & 85\% & 90\% & 95\% & 50\%:95\% \\
    
    \hline
    \multirow{3}{*}{DarkNet-53	}  & \multirow{3}{*}{320} 
    & Precision   & 0.838 & 0.834 & 0.831 & 0.824 & 0.800 & 0.726 & 0.650 & 0.495 & 0.249 & 0.047 & 0.443 \\
    & & Recall    & 0.937 & 0.935 & 0.932 & 0.927 & 0.909 & 0.862 & 0.799 & 0.679 & 0.461 & 0.171 & 0.554 \\
    & & F1-Score  & \textbf{0.884} & \textbf{0.881} & \textbf{0.878} & \textbf{0.872} & \textbf{0.851} & \textbf{0.788} & \textbf{0.716} & \textbf{0.572} & \textbf{0.323} & \textbf{0.073} & \textbf{0.492} \\
    
    \hline
    \multirow{3}{*}{DarkNet-53	}  & \multirow{3}{*}{416} 
    & Precision   & 0.846 & 0.840 & 0.839 & 0.835 & 0.819 & 0.776 & 0.706 & 0.532 & 0.279 & 0.039 & 0.443 \\
    & & Recall    & 0.947 & 0.942 & 0.941 & 0.937 & 0.918 & 0.891 & 0.834 & 0.707 & 0.478 & 0.130 & 0.538 \\
    & & F1-Score  & 0.893 & 0.888 & 0.887 & 0.883 & 0.865 & 0.829 & 0.764 & 0.607 & 0.352 & 0.059 & 0.485 \\

    \hline
    \multirow{3}{*}{DarkNet-53	}  & \multirow{3}{*}{608} 
    & Precision   & 0.841 & 0.835 & 0.829 & 0.821 & 0.800 & 0.773 & 0.713 & 0.555 & 0.229 & 0.026 & 0.433 \\
    & & Recall    & 0.955 & 0.948 & 0.943 & 0.935 & 0.919 & 0.899 & 0.856 & 0.739 & 0.448 & 0.115 &  0.535\\
    & & F1-Score  & \textbf{0.894}* & \textbf{0.887}* & \textbf{0.882}* & \textbf{0.874}* & \textbf{0.855}* & \textbf{0.831}* & \textbf{0.777}* & \textbf{0.633}* & \textbf{0.303}* &  \textbf{0.042}* &  \textbf{0.478}*\\

    \hline
    
\end{tabular}
\end{adjustbox}
\end{center}
\label{tab:YOLO}
\end{table}

\section{Conclusion and future work}
We introduce the TNCR dataset, a new image-based table analysis dataset collected from real images, to aid research in table detection, structure recognition, and classification for document analysis. To evaluate the performance of TNCR, we use the majority of object detection models as a baseline. At each IoU from 50\% to 95\%, models that performed well for table detection were tested. Several combinations were proposed, and the one that performed the best by far was chosen. Table detection is much more difficult than cell structure detection. Experiments show that using deep learning to detect and recognize tables based on images is a promising research direction. We anticipate that the TNCR dataset will unleash the power of deep learning in the table analysis task, while also encouraging more customized network structures to make significant progress.

The Cascade Mask R-CNN, Cascade R-CNN, Cascade RPN, Hybrid Task Cascade (HTC), and YOLO achieve f1 score of 0.844, 0.841, 0.804, 0.840 and 0.492  receptivity.

For future work, Due to the presence of a large amount of tabular data in documents, the structure recognition task is critical in terms of its applicability in business and finance. We intend to expand the dataset by adding more real labeled images. We'll improve a new table detection model to address persistent issues with recognizing structures that are in close proximity to other elements of interest in an image. We Also plan to balance the classes of dataset for classification task.
\bibliographystyle{elsarticle-num} 
\bibliography{elsarticle-template-num}

\begin{thebibliography}{10}
\expandafter\ifx\csname url\endcsname\relax
  \def\url#1{\texttt{#1}}\fi
\expandafter\ifx\csname urlprefix\endcsname\relax\def\urlprefix{URL }\fi
\expandafter\ifx\csname href\endcsname\relax
  \def\href#1#2{#2} \def\path#1{#1}\fi

\bibitem{8270123}
S.~Schreiber, S.~Agne, I.~Wolf, A.~Dengel, S.~Ahmed, Deepdesrt: Deep learning
  for detection and structure recognition of tables in document images, in:
  2017 14th IAPR International Conference on Document Analysis and Recognition
  (ICDAR), Vol.~01, 2017, pp. 1162--1167.
\newblock \href {https://doi.org/10.1109/ICDAR.2017.192}
  {\path{doi:10.1109/ICDAR.2017.192}}.

\bibitem{traquair2019deep}
M.~Traquair, E.~Kara, B.~Kantarci, S.~Khan, Deep learning for the detection of
  tabular information from electronic component datasheets, in: 2019 IEEE
  Symposium on Computers and Communications (ISCC), IEEE, 2019, pp. 1--6.

\bibitem{gilani2017table}
A.~Gilani, S.~R. Qasim, I.~Malik, F.~Shafait, Table detection using deep
  learning, in: 2017 14th IAPR international conference on document analysis
  and recognition (ICDAR), Vol.~1, IEEE, 2017, pp. 771--776.

\bibitem{tran2015table}
D.~N. Tran, T.~A. Tran, A.~Oh, S.~H. Kim, I.~S. Na, Table detection from
  document image using vertical arrangement of text blocks, International
  Journal of Contents 11~(4) (2015) 77--85.

\bibitem{7490132}
L.~Hao, L.~Gao, X.~Yi, Z.~Tang, A table detection method for pdf documents
  based on convolutional neural networks, in: 2016 12th IAPR Workshop on
  Document Analysis Systems (DAS), 2016, pp. 287--292.
\newblock \href {https://doi.org/10.1109/DAS.2016.23}
  {\path{doi:10.1109/DAS.2016.23}}.

\bibitem{mao2003document}
S.~Mao, A.~Rosenfeld, T.~Kanungo, Document structure analysis algorithms: a
  literature survey, in: Document Recognition and Retrieval X, Vol. 5010,
  International Society for Optics and Photonics, 2003, pp. 197--207.

\bibitem{kara2020holistic}
E.~Kara, M.~Traquair, M.~Simsek, B.~Kantarci, S.~Khan, Holistic design for deep
  learning-based discovery of tabular structures in datasheet images,
  Engineering Applications of Artificial Intelligence 90 (2020) 103551.

\bibitem{sarkar2019document}
M.~Sarkar, M.~Aggarwal, A.~Jain, H.~Gupta, B.~Krishnamurthy, Document structure
  extraction for forms using very high resolution semantic segmentation, no.
  February (2019).

\bibitem{lecun1995convolutional}
Y.~LeCun, Y.~Bengio, et~al., Convolutional networks for images, speech, and
  time series, The handbook of brain theory and neural networks 3361~(10)
  (1995) 1995.

\bibitem{zhao2019object}
Z.-Q. Zhao, P.~Zheng, S.-t. Xu, X.~Wu, Object detection with deep learning: A
  review, IEEE transactions on neural networks and learning systems 30~(11)
  (2019) 3212--3232.

\bibitem{lawrence1997face}
S.~Lawrence, C.~L. Giles, A.~C. Tsoi, A.~D. Back, Face recognition: A
  convolutional neural-network approach, IEEE transactions on neural networks
  8~(1) (1997) 98--113.

\bibitem{gehring2017convolutional}
J.~Gehring, M.~Auli, D.~Grangier, D.~Yarats, Y.~N. Dauphin, Convolutional
  sequence to sequence learning, in: International Conference on Machine
  Learning, PMLR, 2017, pp. 1243--1252.

\bibitem{Abdallah_Abdelrahman}
A.~Abdallah, M.~Kasem, M.~A. Hamada, S.~Sdeek,
  \href{https://doi.org/10.1145/3410352.3410744}{Automated question-answer
  medical model based on deep learning technology}, in: Proceedings of the 6th
  International Conference on Engineering \& MIS 2020, ICEMIS'20, Association
  for Computing Machinery, New York, NY, USA, 2020.
\newblock \href {https://doi.org/10.1145/3410352.3410744}
  {\path{doi:10.1145/3410352.3410744}}.
\newline\urlprefix\url{https://doi.org/10.1145/3410352.3410744}

\bibitem{abdel2014convolutional}
O.~Abdel-Hamid, A.-r. Mohamed, H.~Jiang, L.~Deng, G.~Penn, D.~Yu, Convolutional
  neural networks for speech recognition, IEEE/ACM Transactions on audio,
  speech, and language processing 22~(10) (2014) 1533--1545.

\bibitem{paszke2016enet}
A.~Paszke, A.~Chaurasia, S.~Kim, E.~Culurciello, Enet: A deep neural network
  architecture for real-time semantic segmentation, arXiv preprint
  arXiv:1606.02147 (2016).

\bibitem{li2014medical}
Q.~Li, W.~Cai, X.~Wang, Y.~Zhou, D.~D. Feng, M.~Chen, Medical image
  classification with convolutional neural network, in: 2014 13th international
  conference on control automation robotics \& vision (ICARCV), IEEE, 2014, pp.
  844--848.

\bibitem{Abdallah_2020}
A.~Abdallah, M.~Hamada, D.~Nurseitov,
  \href{http://dx.doi.org/10.3390/jimaging6120141}{Attention-based fully gated
  cnn-bgru for russian handwritten text}, Journal of Imaging 6~(12) (2020) 141.
\newblock \href {https://doi.org/10.3390/jimaging6120141}
  {\path{doi:10.3390/jimaging6120141}}.
\newline\urlprefix\url{http://dx.doi.org/10.3390/jimaging6120141}

\bibitem{nurseitov2020hkr}
D.~Nurseitov, K.~Bostanbekov, D.~Kurmankhojayev, A.~Alimova, A.~Abdallah, Hkr
  for handwritten kazakh \& russian database, arXiv preprint arXiv:2007.03579
  (2020).

\bibitem{Daniyar_2020}
G.~A. {Daniyar Nurseitov, Kairat Bostanbekov, Maksat Kanatov, Anel Alimova,
  Abdelrahman Abdallah}, {Classification of Handwritten Names of Cities and
  Handwritten Text Recognition using Various Deep Learning Models}, Advances in
  Science, Technology and Engineering Systems Journal 5~(5) (2020) 934--943.
\newblock \href {https://doi.org/10.25046/aj0505114}
  {\path{doi:10.25046/aj0505114}}.

\bibitem{ren2015faster}
S.~Ren, K.~He, R.~Girshick, J.~Sun, Faster r-cnn: Towards real-time object
  detection with region proposal networks, arXiv preprint arXiv:1506.01497
  (2015).

\bibitem{cai2019cascade}
Z.~Cai, N.~Vasconcelos, Cascade r-cnn: high quality object detection and
  instance segmentation, IEEE transactions on pattern analysis and machine
  intelligence (2019).

\bibitem{He_2017}
K.~He, G.~Gkioxari, P.~Dollar, R.~Girshick, Mask r-cnn, 2017 IEEE International
  Conference on Computer Vision (ICCV) (Oct 2017).

\bibitem{vu2019cascade}
T.~Vu, H.~Jang, T.~X. Pham, C.~D. Yoo, Cascade rpn: Delving into high-quality
  region proposal network with adaptive convolution, in: Conference on Neural
  Information Processing Systems (NeurIPS), 2019.

\bibitem{chen2019hybrid}
K.~Chen, J.~Pang, J.~Wang, Y.~Xiong, X.~Li, S.~Sun, W.~Feng, Z.~Liu, J.~Shi,
  W.~Ouyang, C.~C. Loy, D.~Lin, Hybrid task cascade for instance segmentation,
  in: IEEE Conference on Computer Vision and Pattern Recognition, 2019.

\bibitem{SunXLW19}
K.~Sun, B.~Xiao, D.~Liu, J.~Wang, Deep high-resolution representation learning
  for human pose estimation, in: CVPR, 2019.

\bibitem{SunZJCXLMWLW19}
K.~Sun, Y.~Zhao, B.~Jiang, T.~Cheng, B.~Xiao, D.~Liu, Y.~Mu, X.~Wang, W.~Liu,
  J.~Wang, High-resolution representations for labeling pixels and regions,
  CoRR abs/1904.04514 (2019).

\bibitem{zhang2020resnest}
H.~Zhang, C.~Wu, Z.~Zhang, Y.~Zhu, Z.~Zhang, H.~Lin, Y.~Sun, T.~He, J.~Muller,
  R.~Manmatha, M.~Li, A.~Smola, Resnest: Split-attention networks, arXiv
  preprint arXiv:2004.08955 (2020).

\bibitem{redmon2018yolov3}
J.~Redmon, A.~Farhadi, Yolov3: An incremental improvement (2018).
\newblock \href {http://arxiv.org/abs/1804.02767} {\path{arXiv:1804.02767}}.

\bibitem{DynamicRCNN}
H.~Zhang, H.~Chang, B.~Ma, N.~Wang, X.~Chen, Dynamic {R-CNN}: Towards high
  quality object detection via dynamic training, arXiv preprint
  arXiv:2004.06002 (2020).

\bibitem{he2016deep}
K.~He, X.~Zhang, S.~Ren, J.~Sun, Deep residual learning for image recognition,
  in: Proceedings of the IEEE conference on computer vision and pattern
  recognition, 2016, pp. 770--778.

\bibitem{xie2017aggregated}
S.~Xie, R.~Girshick, P.~Doll{\'a}r, Z.~Tu, K.~He, Aggregated residual
  transformations for deep neural networks, in: Proceedings of the IEEE
  conference on computer vision and pattern recognition, 2017, pp. 1492--1500.

\bibitem{gobel2013icdar}
M.~G{\"o}bel, T.~Hassan, E.~Oro, G.~Orsi, Icdar 2013 table competition, in:
  2013 12th International Conference on Document Analysis and Recognition,
  IEEE, 2013, pp. 1449--1453.

\bibitem{shahab2010open}
A.~Shahab, F.~Shafait, T.~Kieninger, A.~Dengel, An open approach towards the
  benchmarking of table structure recognition systems, in: Proceedings of the
  9th IAPR International Workshop on Document Analysis Systems, 2010, pp.
  113--120.

\bibitem{fang2012dataset}
J.~Fang, X.~Tao, Z.~Tang, R.~Qiu, Y.~Liu, Dataset, ground-truth and performance
  metrics for table detection evaluation, in: 2012 10th IAPR International
  Workshop on Document Analysis Systems, IEEE, 2012, pp. 445--449.

\bibitem{siegel2018extracting}
N.~Siegel, N.~Lourie, R.~Power, W.~Ammar, Extracting scientific figures with
  distantly supervised neural networks, in: Proceedings of the 18th ACM/IEEE on
  joint conference on digital libraries, 2018, pp. 223--232.

\bibitem{li2020tablebank}
M.~Li, L.~Cui, S.~Huang, F.~Wei, M.~Zhou, Z.~Li, Tablebank: Table benchmark for
  image-based table detection and recognition, in: Proceedings of the 12th
  Language Resources and Evaluation Conference, 2020, pp. 1918--1925.

\bibitem{dejean_herve_2019_2649217}
H.~Déjean, J.-L. Meunier, L.~Gao, Y.~Huang, Y.~Fang, F.~Kleber, E.-M. Lang,
  \href{https://doi.org/10.5281/zenodo.2649217}{{ICDAR 2019 Competition on
  Table Detection and Recognition (cTDaR)}}, http://sac.founderit.com/ (Apr.
  2019).
\newblock \href {https://doi.org/10.5281/zenodo.2649217}
  {\path{doi:10.5281/zenodo.2649217}}.
\newline\urlprefix\url{https://doi.org/10.5281/zenodo.2649217}

\bibitem{itonori1993table}
K.~Itonori, Table structure recognition based on textblock arrangement and
  ruled line position, in: Proceedings of 2nd International Conference on
  Document Analysis and Recognition (ICDAR'93), IEEE, 1993, pp. 765--768.

\bibitem{seo2015junction}
W.~Seo, H.~I. Koo, N.~I. Cho, Junction-based table detection in camera-captured
  document images, International Journal on Document Analysis and Recognition
  (IJDAR) 18~(1) (2015) 47--57.

\bibitem{chandran1993structural}
S.~Chandran, R.~Kasturi, Structural recognition of tabulated data, in:
  Proceedings of 2nd International Conference on Document Analysis and
  Recognition (ICDAR'93), IEEE, 1993, pp. 516--519.

\bibitem{inproceedings}
T.~Kieninger, A.~Dengel, The t-recs table recognition and analysis system, Vol.
  1655, 1998, pp. 255--269.

\bibitem{cesarini2002trainable}
F.~Cesarini, S.~Marinai, L.~Sarti, G.~Soda, Trainable table location in
  document images, in: Object recognition supported by user interaction for
  service robots, Vol.~3, IEEE, 2002, pp. 236--240.

\bibitem{6628801}
T.~Kasar, P.~Barlas, S.~Adam, C.~Chatelain, T.~Paquet, Learning to detect
  tables in scanned document images using line information, in: 2013 12th
  International Conference on Document Analysis and Recognition, 2013, pp.
  1185--1189.
\newblock \href {https://doi.org/10.1109/ICDAR.2013.240}
  {\path{doi:10.1109/ICDAR.2013.240}}.

\bibitem{e2009learning}
A.~C. e~Silva, Learning rich hidden markov models in document analysis: Table
  location, in: 2009 10th International Conference on Document Analysis and
  Recognition, IEEE, 2009, pp. 843--847.

\bibitem{kara2019deep}
E.~Kara, M.~Traquair, B.~Kantarci, S.~Khan, Deep learning for recognizing the
  anatomy of tables on datasheets, in: 2019 IEEE Symposium on Computers and
  Communications (ISCC), IEEE, 2019, pp. 1--6.

\bibitem{arif2018table}
S.~Arif, F.~Shafait, Table detection in document images using foreground and
  background features, in: 2018 Digital Image Computing: Techniques and
  Applications (DICTA), IEEE, 2018, pp. 1--8.

\bibitem{8540832}
S.~A. Siddiqui, M.~I. Malik, S.~Agne, A.~Dengel, S.~Ahmed, Decnt: Deep
  deformable cnn for table detection, IEEE Access 6 (2018) 74151--74161.
\newblock \href {https://doi.org/10.1109/ACCESS.2018.2880211}
  {\path{doi:10.1109/ACCESS.2018.2880211}}.

\bibitem{prasad2020cascadetabnet}
D.~Prasad, A.~Gadpal, K.~Kapadni, M.~Visave, K.~Sultanpure, Cascadetabnet: An
  approach for end to end table detection and structure recognition from
  image-based documents, in: Proceedings of the IEEE/CVF Conference on Computer
  Vision and Pattern Recognition Workshops, 2020, pp. 572--573.

\bibitem{long2015fully}
J.~Long, E.~Shelhamer, T.~Darrell, Fully convolutional networks for semantic
  segmentation, in: Proceedings of the IEEE conference on computer vision and
  pattern recognition, 2015, pp. 3431--3440.

\bibitem{everingham2010pascal}
M.~Everingham, L.~Van~Gool, C.~K. Williams, J.~Winn, A.~Zisserman, The pascal
  visual object classes (voc) challenge, International journal of computer
  vision 88~(2) (2010) 303--338.

\bibitem{paliwal2019tablenet}
S.~S. Paliwal, D.~Vishwanath, R.~Rahul, M.~Sharma, L.~Vig, Tablenet: Deep
  learning model for end-to-end table detection and tabular data extraction
  from scanned document images, in: 2019 International Conference on Document
  Analysis and Recognition (ICDAR), IEEE, 2019, pp. 128--133.

\bibitem{kavasidis2018saliency}
I.~Kavasidis, S.~Palazzo, C.~Spampinato, C.~Pino, D.~Giordano, D.~Giuffrida,
  P.~Messina, A saliency-based convolutional neural network for table and chart
  detection in digitized documents, arXiv preprint arXiv:1804.06236 (2018).

\bibitem{jiang2021tabcellnet}
J.~Jiang, M.~Simsek, B.~Kantarci, S.~Khan, Tabcellnet: Deep learning-based
  tabular cell structure detection, Neurocomputing 440 (2021) 12--23.

\bibitem{zhong2019image}
X.~Zhong, E.~ShafieiBavani, A.~J. Yepes, Image-based table recognition: data,
  model, and evaluation, arXiv preprint arXiv:1911.10683 (2019).

\bibitem{luo2021deep}
S.~Luo, M.~Wu, Y.~Gong, W.~Zhou, J.~Poon, Deep structured feature networks for
  table detection and tabular data extraction from scanned financial document
  images, arXiv preprint arXiv:2102.10287 (2021).

\bibitem{chen2019mmdetection}
K.~Chen, J.~Wang, J.~Pang, Y.~Cao, Y.~Xiong, X.~Li, S.~Sun, W.~Feng, Z.~Liu,
  J.~Xu, et~al., Mmdetection: Open mmlab detection toolbox and benchmark, arXiv
  preprint arXiv:1906.07155 (2019).

\end{thebibliography}

\label{Conclusion}
\end{document}